\documentclass[letterpaper,10pt,conference]{ieeeconf}

\IEEEoverridecommandlockouts
\usepackage{amsmath,amssymb}
\DeclareMathSizes{5.5}{5.5}{5}{5}
\usepackage{cite}
\usepackage{graphicx}
\usepackage{float}
\usepackage{stfloats}
\usepackage[table]{xcolor}
\usepackage[hidelinks]{hyperref}
\graphicspath{{./}}

\newcommand{\pendingexp}{\texttt{[PENDING EXPERIMENT]}}
\newcommand{\toverify}{\texttt{[TO VERIFY]}}

\newcommand{\tbd}{\texttt{[TBD]}}

\begin{document}

\title{\LARGE \bf
TACROSS: An Efficient and Low-Cost Scalable Human Touch System Across Heterogeneous Tactile Sensors for Dexterous Robot Learning}
\author{Bo Chen$^{1,6,*}$, Huanzhang Hu$^{2,6,*}$, Junyang Ma$^{5}$, Bo Yue$^{2}$, Fangdi Yu$^{3}$,\\
Haijier Chen$^{4,6}$, Xianxin Lai$^{1,6}$, Shuyu Pan$^{6}$, Zhen Yang$^{1,6}$, Xiaoquan Sun$^{1,6}$,\\
Wenze Cui$^{1}$, Zhongliang Jiang$^{1}$, Shaopeng Liu$^{1,\dagger}$, and Jiayu Chen$^{1,6,\dagger}$\\[-0.1em]
\scriptsize $^{1}$The University of Hong Kong $^{2}$The Chinese University of Hong Kong (Shenzhen) $^{3}$Ocean University of China\\[-0.1em]
\scriptsize $^{4}$Wuhan University $^{5}$Wuhan Textile University $^{6}$INFIFORCE\\[-0.1em]
\scriptsize $^{*}$Equal contribution; $^{\dagger}$Corresponding author}
\IEEEaftertitletext{%
\vspace{-0.35\baselineskip}%
\begin{minipage}{\textwidth}
\centering
\includegraphics[width=\textwidth]{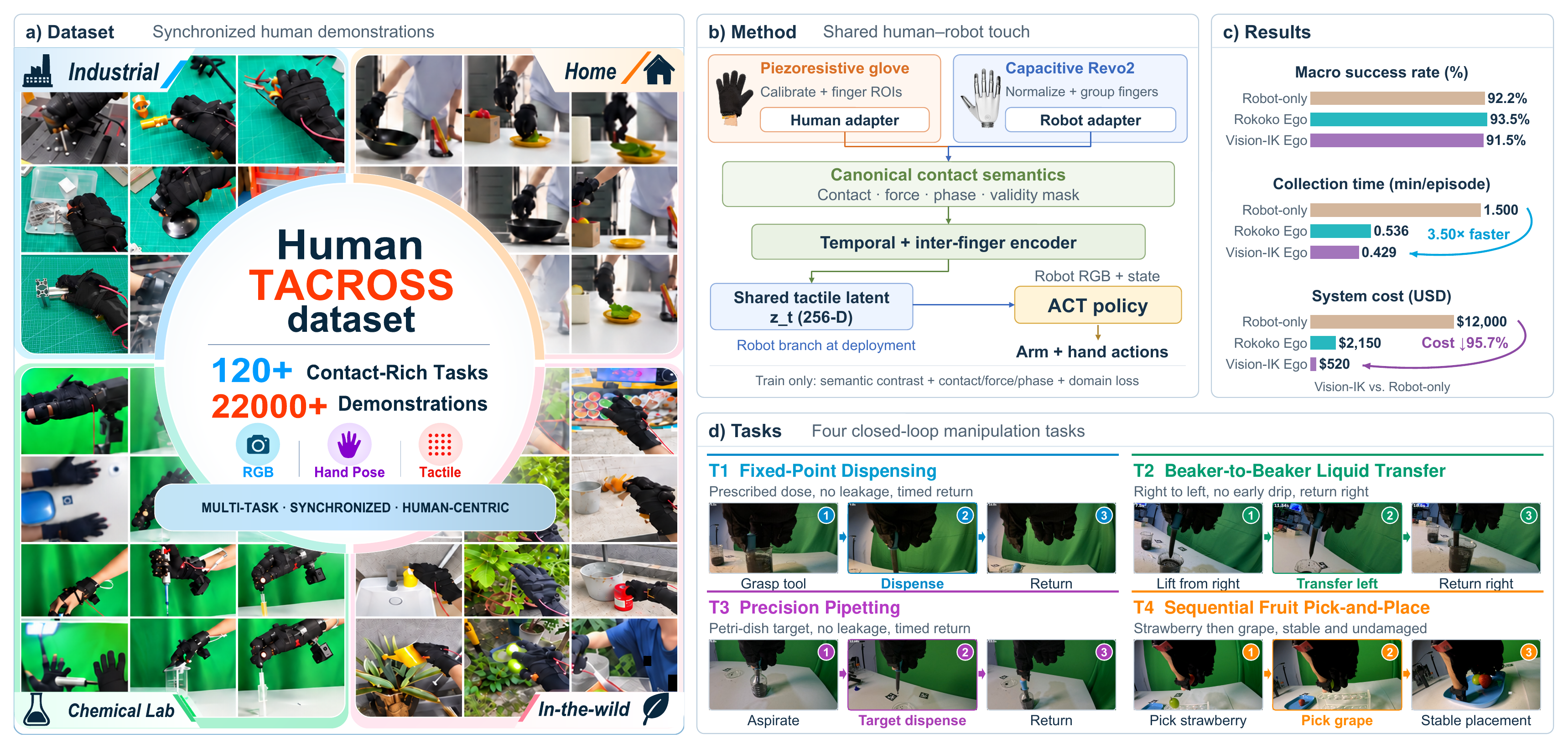}\\[-0.15em]
\refstepcounter{figure}%
\parbox{\textwidth}{\footnotesize Fig.~\thefigure.\quad TACROSS overview. (a) Synchronized human demonstrations with egocentric RGB, hand pose, and tactile observations. (b) Contact canonicalization and sensor-specific adaptation align piezoresistive glove and capacitive Revo2 observations. Shared temporal and inter-finger encoding produces tactile latents for robot-supervised ACT learning. Training uses masked contact-semantic targets and contrastive alignment, while deployment uses robot observations only. (c) Vision-IK Ego achieves similar success (91.5\% vs. 92.2\%) at $3.5\times$ the average collection rate and 95.7\% lower equipment cost than Robot-only. (d) Four contact-rich manipulation tasks for closed-loop evaluation.}
\label{fig:overview}
\end{minipage}%
\vspace{0.45\baselineskip}%
}
\maketitle

\begin{abstract}
Collecting tactile demonstrations on robots is costly and slow, motivating the use of lower-cost human tactile gloves for scalable data collection. However, human capacitive/piezoresistive gloves and robotic tactile sensors differ fundamentally in transduction principle, sensor layout, spatial resolution, and dynamic response, making alignment of raw sensor channels ill-posed. To address this problem, we present TACROSS, a scalable system for learning from human touch and transferring it to robots that bridges this heterogeneity by aligning tactile streams at the level of contact events rather than raw sensor values. The hardware component of TACROSS integrates a piezoresistive glove with five layers and a cost of \$10.86 with 285 sensing points. To align contact semantics, we design canonicalizers and residual adapters that map heterogeneous signals into a shared tactile latent with 256 dimensions via a temporal Transformer with attention across fingers. We further introduce a robot-grounded policy learning scheme in which robot demonstrations provide the sole source of ground-truth action supervision, while human demonstrations support tactile representation learning and provide confidence-weighted auxiliary supervision through valid retargeted hand targets. We evaluate our system on four contact-rich manipulation tasks. Compared to conventional teleoperation, our proposed system achieves a 3.5$\times$ efficiency improvement while reducing demonstration acquisition equipment cost by 95.7\%. We will open-source the TACROSS hardware and software system and publicly release a tactile dataset comprising over 150 hours of recordings. Project page: \href{https://tacross-touch-project.github.io/}{\textcolor{blue}{\nolinkurl{https://tacross-touch-project.github.io/}}}.
\end{abstract}

\section{Introduction}

Force tactile interaction technology, an advanced human--computer interaction modality that mimics the sense of physical touch, is increasingly serving as a bridge between the digital and physical worlds~\cite{hu2025forcetactile}. Contact-rich dexterous manipulation depends on events, including grasp onset, loading, stability, slip, and release, that RGB observation leaves visually ambiguous~\cite{forcemimic2025}. A central question is how to leverage human tactile demonstrations collected efficiently at low cost to reduce robot data collection demands while preserving effective manipulation performance and retaining robot action supervision~\cite{humanego2026}. Robot tactile demonstrations expose these events directly~\cite{tactilealoha2025}, but every collection minute occupies the target arm, hand, and sensor platform, incurring operation, reset, and calibration costs. Under our accounting, robot teleoperation records only 40 episodes/h on a USD~12,000 system. Scalable tactile gloves capture distributed human contact patterns~\cite{stag2019,opentouch2025,htbench2026}, while systems that collect visual and tactile data without robots demonstrate a practical route to acquiring contact-rich manipulation data~\cite{freetacman2025}. Human tactile gloves offer a decoupled, faster, and cheaper alternative: our USD~2,150 Ego-R and USD~520 Ego-V systems reach 112 and 140 episodes/h, corresponding to 2.80$\times$ and 3.50$\times$ the robot rate, together yielding 159.09~h of recorded human trajectories, 23,677 recorded demonstrations, and 21,412 valid ones (Table~\ref{tab:acquisition-rebuild}). 

Yet this larger human corpus is usable only if it can be related to robot touch. UniTouch and AnyTouch learn shared representations across heterogeneous vision-based tactile sensors~\cite{unitouch2024,anytouch2025}, motivating our study of alignment across different transduction principles. The human glove is piezoresistive, whereas Revo2 sensing is capacitive. The two differ in transduction, taxel layout, spatial support, resolution, numerical scale, noise, drift, hysteresis, and temporal response. Equal channel values therefore need not denote equal contact, and identical physical events can activate different numbers and locations of taxels. Direct matching of individual channels is thus ill-posed: sensor identity and acquisition artifacts can obscure the interaction state required for control. We term this mismatch the \textbf{Sensing Gap}.

EgoMimic combines human and robot demonstrations through alignment across domains and joint training~\cite{egomimic2024}, while DexPilot maps observed human hand motion to a dexterous robot hand--arm system~\cite{dexpilot2020}. Even if the sensing gap were bridged at the representation level, a second mismatch remains at the action level: human hand motion, however precisely retargeted, is not an action executed by Revo2. Kinematic and morphological differences between the human hand and the robotic end-effector mean that pseudo actions derived from human demonstrations carry retargeting error and cannot be treated as ground truth. Treating these pseudo-labels as ground-truth robot actions can introduce retargeting errors into the learned policy. We term this mismatch the \textbf{Action Gap}. We address the Action Gap by grounding policy learning in robot demonstrations.

We present \textbf{TACROSS}, an integrated hardware--software system for human tactile data acquisition and cross-sensor transfer for robot learning. To solve the Sensing Gap, we propose canonicalizers with validity masks together with residual adapters and a shared causal temporal encoder with attention across fingers. The canonicalizers convert raw human and robot signals into a common masked anatomical contact field, capturing active region, contact probability, coarse force, contact area, duration, loading dynamics, phase, instability, and release, with explicit validity masks distinguishing missing measurements from measured zero contact. The residual adapters account for differences in scale, noise, drift, and hysteresis between domains. And the shared encoder represents contact event dynamics and coordination among fingers across embodiments, together yielding a shared tactile latent with 256 dimensions. To address the Action Gap, we ground policy learning in executed robot actions, which provide the sole source of ground-truth action supervision. Retargeted human hand targets serve as pseudo-labels for auxiliary supervision, weighted by confidence and masked by validity.

The contributions of this work are:
\begin{itemize}
    \item \textbf{Efficient and Low-Cost Data Acquisition System.} We present \textbf{TACROSS}, a hardware--software system with low-cost tactile acquisition, and evaluate its collection throughput and downstream robot performance.
    \item \textbf{Tactile Alignment across Sensors and Policy Learning from Robot Demonstrations.} We align heterogeneous human and robot touch in a shared contact-semantic space with validity masks and ground policy learning in executed robot actions.
    \item \textbf{Open-Source Resources and Tactile Dataset.} We will open-source the tactile glove designs, robot platform configurations, and software for data logging, processing, and model training, and release 183.95~h of recorded trajectories: 159.09~h of human data and 24.86~h of robot data.
\end{itemize}

All design files are available at the \href{https://tacross-touch-project.github.io/}{\textcolor{blue}{project website}}.

\section{Related Work}

\subsection{Human Tactile Demonstration Acquisition}
STAG and Flex-Glove enable distributed human tactile capture through inexpensive sensing and personalized fabrication~\cite{stag2019,flexglove2025}. OSMO uses the same glove on humans and robots, while DexViTac combines multimodal acquisition and kinematics-grounded representation learning with matched tactile hardware~\cite{osmo2025,dexvitac2026}. ForceMimic records interaction wrenches and motion for force-centric imitation learning. TACROSS connects a piezoresistive human glove to a capacitive robot hand, addressing sensor heterogeneity with measurement validity and causal synchronization.

\subsection{Cross-Sensor Tactile Alignment}
Domain-adversarial training uses gradient reversal to learn task-discriminative, domain-invariant features~\cite{dann2016}. UniTacHand aligns heterogeneous touch through masked MANO-UV maps and paired interactions with contrastive, reconstruction, and adversarial objectives~\cite{unitachand2025}. TactAlign learns a rectified-flow mapping from interaction-derived pseudo pairs, supporting unpaired demonstrations and dynamic contact~\cite{tactalign2026}. TACROSS represents regional contact state, relative intensity, duration, spatial descriptors, and phase. Sensor-specific adapters feed shared causal temporal and inter-finger encoders. Positive windows match task/object metadata and contact semantics without cross-domain frame synchronization. Field-level validity masks distinguish missing measurements from valid no-contact observations.

\subsection{Learning from Human and Robot Demonstrations}
EgoMimic jointly trains policies on human and robot demonstrations~\cite{egomimic2024}. FreeTacMan combines temporal tactile pretraining with ACT and evaluates optical cross-sensor generalization~\cite{freetacman2025}. MimicTouch learns from multimodal human demonstrations and bridges the embodiment gap through online residual reinforcement learning~\cite{mimictouch2025}. TACROSS uses human touch for representation learning and robot actions for ground-truth supervision. HaMeR~\cite{hamer2024} and calibrated retargeting provide human hand pseudo-targets for confidence-weighted, validity-masked auxiliary supervision. Arm targets require separate supervision. ACT~\cite{act2023} takes the robot tactile latent, RGB, and proprioception, using robot observations at deployment.

\section{TACROSS System and Method}
\subsection{Problem Formulation and Overview}
TACROSS consists of a human tactile acquisition system, a contact-semantic encoder with human and robot input branches, and an ACT policy. Training uses human and robot episodes, deployment uses the robot branch and policy.

Human and robot episodes are defined as
\begin{align}
\mathcal{D}_{H}&=\{I_t^H,q_t^H,X_t^H,\tilde a_t^{R,H},w_t^H,\tau_t\}_{t=1}^{T_H},\\
\mathcal{D}_{R}&=\{I_t^R,q_t^R,X_t^R,a_t^R,\tau_t\}_{t=1}^{T_R},
\end{align}
where $I$, $q$, $X$, and $\tau$ denote RGB images, kinematics/proprioception, tactile observations, and native timestamps. The executed robot action $a_t^R$ provides the ground-truth action label. The optional retargeted human hand target $\tilde a_t^{R,H}$ serves as a pseudo-label for auxiliary supervision, with $w_t^H$ encoding confidence and validity.

For each domain, a canonicalizer produces regional contact fields and validity masks. An adapter for each domain converts these fields into tokens, which are processed by shared encoders for temporal dynamics and finger interactions. The complete tactile path is denoted by
\begin{equation}
z_t^e=f_e(X_{1:t}^e,q_{1:t}^e),\qquad e\in\{H,R\},
\end{equation}
The robot policy combines $z_t^R$ with RGB and proprioception to predict an action chunk:
\begin{equation}
a_{t:t+H}^R=\pi(I_t^R,q_t^R,z_t^R).
\end{equation}
Sections B--D describe acquisition, tactile encoding, and policy training, respectively.

\subsection{Human Tactile Acquisition}
\begin{figure}[!ht]
\centering
\includegraphics[width=0.95\columnwidth]{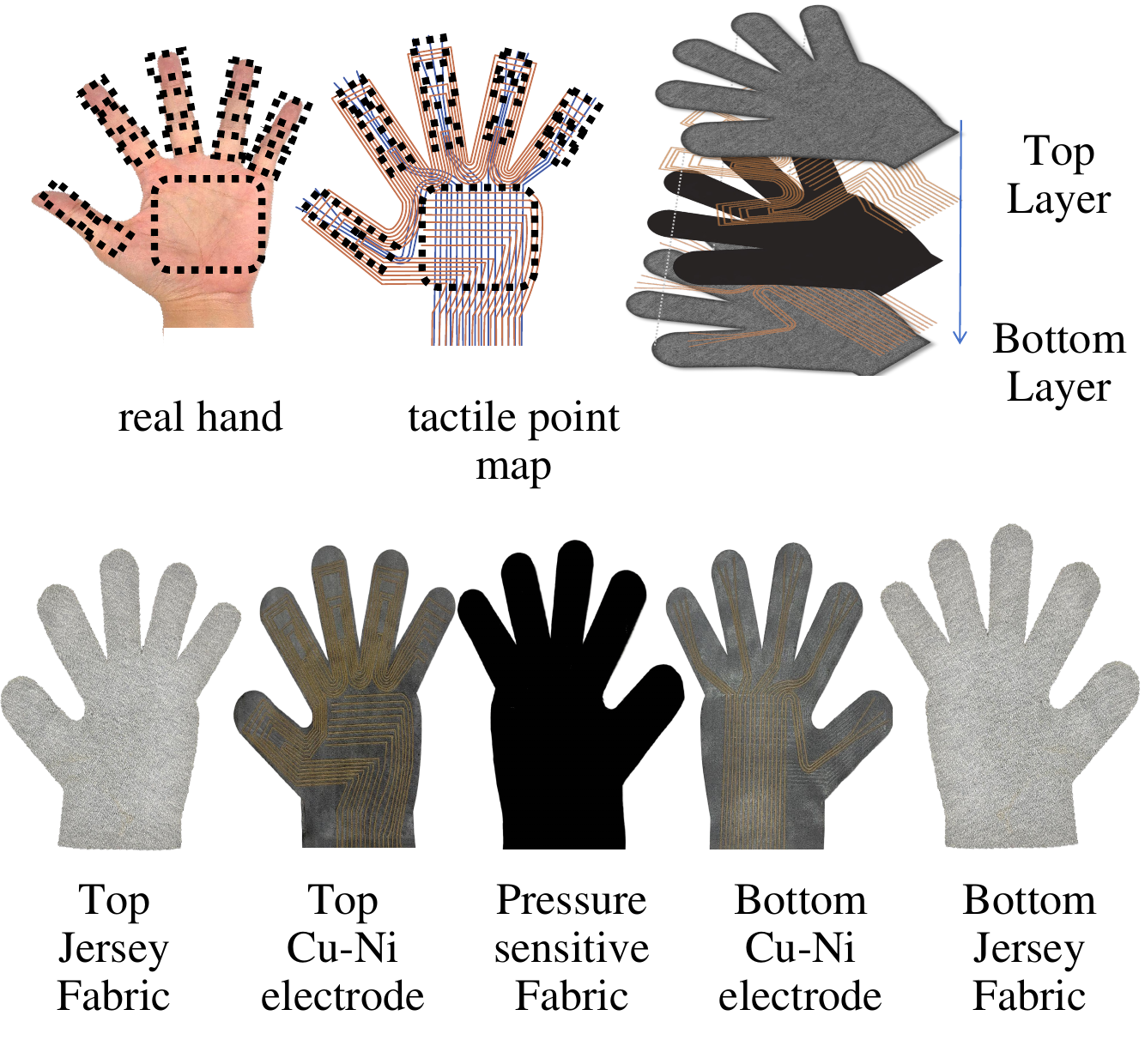}
\caption{TACROSS glove: sensing coverage,and five-layer structure..}
\label{fig:glove-structure}
\end{figure}

\textbf{Glove Design and Readout.}
The glove comprises five layers, including two patterned copper--nickel electrodes, a pressure-sensitive layer, and palmar/dorsal fabric encapsulation (\hyperref[fig:glove-structure]{\textcolor{blue!60!black}{Fig.~\ref*{fig:glove-structure}}}). Its $12\times15$ palm and five $3\times7$ finger arrays provide 285 physical sensing locations. Matrix multiplexing, a 12-bit ADC, and a wrist controller stream frames of 256 values over serial/BLE at approximately 150~Hz. Physical locations and streamed values are distinct, defective channels retain invalid flags. \hyperref[tab:hardware-characterization]{\textcolor{blue!60!black}{Table~\ref*{tab:hardware-characterization}}} reports hardware measurements.

\begin{figure*}[t]
\centering
\includegraphics[width=\textwidth]{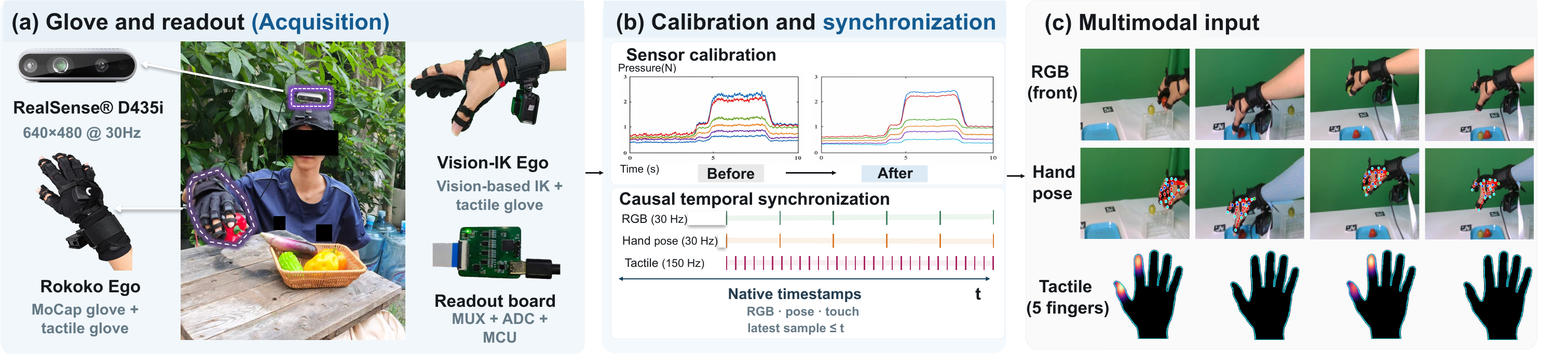}
\caption{TACROSS human demonstration acquisition pipeline. (a) Tactile glove and on-wrist readout integrated with motion capture (Rokoko Ego) or vision-based inverse kinematics (Vision-IK Ego). (b) Sensor calibration and causal synchronization using native timestamps, selecting the latest available sample at or before each reference time. (c) Aligned RGB images, hand poses, and five-finger tactile observations form multimodal episodes for subsequent tactile representation and policy learning.}
\label{fig:hardware-to-learning}
\end{figure*}

\textbf{Calibration and Causal Synchronization.}
For channel $i$, $x_{i,t}$ is the raw reading, $b_i$ its unloaded reference, $g_i$ the calibration gain converting baseline-subtracted readings to the calibrated scale, and $m_i\in\{0,1\}$ the channel-validity flag:
\begin{equation}
\widetilde x_{i,t}=m_i\operatorname{clip}(g_i(x_{i,t}-b_i),0,x_{\max}).
\end{equation}
Here, $\operatorname{clip}(y,0,x_{\max})=\min\{x_{\max},\max(0,y)\}$, where $x_{\max}>0$ is the upper bound on that scale. $m_i=0$ suppresses an invalid channel, not a valid no-contact observation.
At time $t$, each modality selects its latest sample with timestamp $\tau\leq t$. Episodes store RGB, pose, raw/calibrated touch, timestamps, sample ages, masks, and valid pseudo actions. \hyperref[fig:hardware-to-learning]{\textcolor{blue!60!black}{Figure~\ref*{fig:hardware-to-learning}}} summarizes the pipeline for acquisition and data conditioning.

\begin{table}[!htbp]
\centering
\caption{Hardware specifications and characterization of the TACROSS glove. F.S. denotes full scale.}
\label{tab:hardware-characterization}
\footnotesize
\setlength{\tabcolsep}{3pt}
\renewcommand{\arraystretch}{1.04}
\begin{tabular}{@{}p{0.29\columnwidth}p{0.66\columnwidth}@{}}
\hline
\textbf{Characteristic} & \textbf{Reported value}\\
\hline
BOM cost & USD 10.86/glove (controller 6.58, materials 4.28)\\
Derived cost/location & USD 0.038 (BOM / 285 locations)\\
Layout and coverage & 285 locations, $\sim$200 cm$^2$, $>85\%$ coverage\\
Frame and rate & 256 values/frame, $\sim$150 Hz\\
Serial timing & 1.72 ms/frame (theoretical)\\
Force test & 0--20 N, max 83.91/255 ADC, no saturation\\
Repeatability & 1.38\% F.S. (loaded)\\
Nonlinearity & 8.26\% F.S., $R^2=0.90$\\
Hysteresis & 11.65\% F.S. (taxel 12, 7)\\
Drift & 0.39--0.41\% F.S. / 10 min\\
Mass and wearability & Glove $<50$ g, system $<85$ g, donning $<30$ s, wear $>2$ h\\
\hline
\end{tabular}
\end{table}

\begin{figure*}[t]
\centering
\includegraphics[width=\textwidth]{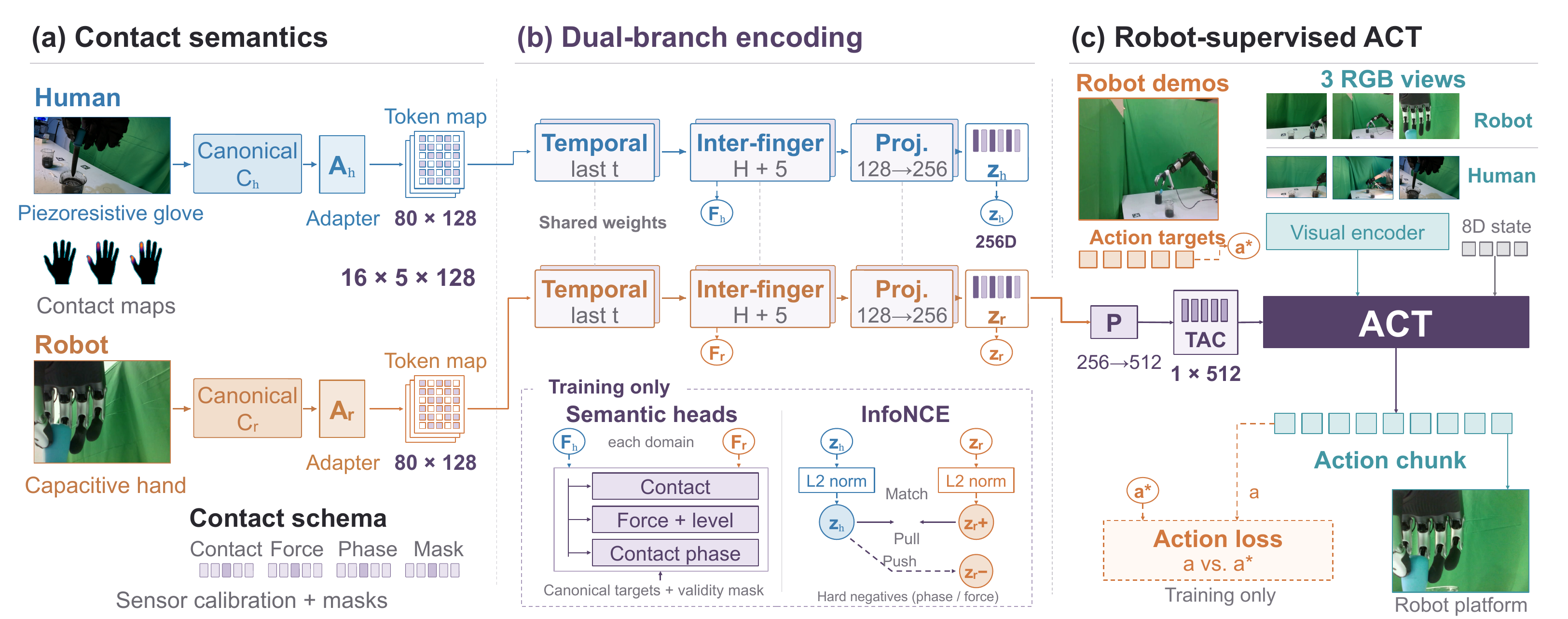}
\caption{Contact canonicalization and adaptation, shared spatiotemporal encoding, and ACT learning. Each window contains 80 tokens (16 frames $\times$ five fingers), processed by shared Transformers with two layers, four attention heads, and features of dimension 128. $H$ is a learned hand token, $F$ denotes finger features, and repeated labels identify the same tensor. Training uses masked canonical targets in both domains and contrastive attraction/repulsion with phase/force hard negatives. $P$ projects latents from 256 dimensions to tokens with 512 dimensions through Linear, GELU, and LayerNorm layers. The illustrated state has eight dimensions and uses one degree of freedom for hand control, Sec.~\ref{sec:robot-grounding} defines the articulated action interface with 13 dimensions. Deployment uses robot observations.}
\label{fig:cross-sensor-alignment}
\end{figure*}

\subsection{Contact-Semantic Alignment}
Calibration standardizes part of the numerical preprocessing, but sensor layout, contact distribution, and dynamic response still differ between human and robot hands. TACROSS aligns regional contact semantics (\hyperref[fig:cross-sensor-alignment]{\textcolor{blue!60!black}{Fig.~\ref*{fig:cross-sensor-alignment}}}), reducing domain differences irrelevant to control while retaining contact information.

\subsubsection{Contact Canonicalization across Embodiments}
For shared anatomical region $r$, $g_e$ aggregates the available measurements in domain $e$. The domain-specific configuration $\psi_e$ specifies the channel-to-region correspondence and calibration settings. Correspondence follows anatomy rather than channel number. The canonicalizers are
\begin{equation}
(C_t^e,M_t^e)=g_e(X_{1:t}^e,q_{1:t}^e,\psi_e),\quad e\in\{H,R\}.
\end{equation}
Here, $C_t^e$ stacks regional field vectors $c_{t,r}^e$, and $M_t^e$ stacks their componentwise validity masks $m_{t,r}^e$. Both domains use the same field order:
\begin{equation}
c_{t,r}^e=[p,\widetilde F,\Delta\widetilde F,d,A,u,v,\rho,\gamma]_{t,r}^e,
\end{equation}
encoding contact probability, relative intensity/change, duration, area, center of pressure, instability proxy, and calibration confidence. Four intensity levels and six causal phases (no contact, onset, loading, stable, instability candidate, release) describe contact evolution. Only available fields are comparable: an unsupported field or region has zero value and mask, whereas valid readings without contact retain valid masks. Here, $\widetilde F$ encodes relative tactile intensity rather than calibrated absolute force, and $\rho$ is an instability proxy rather than a direct slip measurement.

\subsubsection{Adapters for Each Sensor}
The scale, response statistics, and temporal behavior of common fields still depend on the sensor. Separate shallow adapters $A_H,A_R$ map them to regional tokens with a common interface:
\begin{equation}
u_{t,r}^e=A_e([c_{t,r}^e\odot m_{t,r}^e,m_{t,r}^e,e^F,e^P,x_{t,r},e_r]).
\end{equation}
The inputs are masked fields, masks, intensity/phase embeddings, regional context, and anatomical identity. Domain identity selects the adapter, it is not an additional input to the shared encoder.

\subsubsection{Shared Spatiotemporal Encoding}
A causal window of 16 frames at 15~Hz enters temporal and finger Transformers, each with two layers. Temporal attention models contact evolution, attention across fingers models coordination. Both domains share encoder weights, with padding masks for missing fingers. A learned hand token yields a latent $z_t^e$ with 256 dimensions, finger tokens feed contact, force, and phase heads. Hand latents are normalized for contrastive learning.

\subsubsection{Objectives for Alignment across Domains}
Positive pairs of human and robot windows are selected for semantic compatibility using task/object, primitive, metadata on active fingers, intensity levels, and phase, without requiring synchronized frames across domains. Multi-positive contrastive learning attracts compatible hand latents and separates negatives. Supervision for contact, force regression, ordinal force, and phase retain predictive contact information, together with contrastive discrimination, they discourage a constant representation. Temporal regularization constrains evolution, while domain-adversarial learning discourages sensor identity~\cite{dann2016}. The combined objective is
\begingroup
\setlength{\abovedisplayskip}{6pt}
\setlength{\belowdisplayskip}{6pt}
\setlength{\abovedisplayshortskip}{4pt}
\setlength{\belowdisplayshortskip}{6pt}
\begin{equation}
\begin{aligned}
\mathcal L_{\mathrm{rep}} &= \sum_{j=1}^{7}\lambda_j\mathcal L_j,\\[2pt]
(\lambda_j)_{j=1}^{7} &= (1,1,0.5,0.5,1,0.2,0.1).
\end{aligned}
\end{equation}
\endgroup
The terms $\mathcal L_1$--$\mathcal L_7$ are, respectively, the contrastive loss on hand latents, contact-prediction loss, relative-intensity regression loss, intensity-level loss, phase-classification loss, temporal regularizer, and domain-adversarial loss.

\subsection{Robot-Grounded Policy Learning}
\label{sec:robot-grounding}
\textbf{Robot Demonstrations and Retargeted Human Actions.}
Robot demonstrations provide the sole source of ground-truth action supervision. Human hand targets generated by HaMeR and a calibrated geometric retargeter serve as pseudo-labels for confidence-weighted auxiliary supervision with validity masking. From RGB, HaMeR recovers 21 hand keypoints in three dimensions. We express them in a palm frame centered at the wrist and normalize by palm width. The direction from the index MCP to the little finger MCP defines the lateral axis. The direction from the wrist to the middle finger MCP and cross products complete the orthonormal frame.

For non-thumb finger $f$, let $j\in\{\mathrm{MCP,PIP,DIP}\}$ index its flexion joints. The calibrated angle $\hat\theta_{f,j}$ measures joint closure relative to the user's open/fist references. Whole-finger closure is $c_f=\sum_j\beta_{f,j}\hat\theta_{f,j}$, where $\beta_{f,j}\geq0$ is the contribution of joint $j$ and $\sum_j\beta_{f,j}=1$.
 Thumb flexion, abduction, and the relative position of the thumb and index fingertips describe opposition. These features map to Revo2's six active control channels according to its actuation and coupling geometry. User calibration poses and robot safe endpoints determine channel offsets, scales, and directions. Low-pass filtering, limits on changes per cycle, and joint bounds precede SDK execution. Invalid poses trigger a brief hold, followed by the configured safe state for sustained loss, invalid pseudo targets are excluded from supervision.

\textbf{Policy Interface and Training.}~\cite{atomvla2026,song2026practicality,himemwam2026}
We use two modes of hand control. In \emph{single-DOF mode}, the entire hand is treated as one control unit with only open and close commands, $b\in\{0,1\}$. In \emph{six-DOF retargeting mode}, the geometric mapping above produces targets $h\in\mathbb R^6$ for Revo2's six active joints. With targets for seven arm joints, the corresponding action interfaces are
\begin{equation}
a^{(8)}=[a^{\mathrm{arm}},b],\qquad a^{(13)}=[a^{\mathrm{arm}},h].
\end{equation}
ACT learns from ground-truth robot actions and receives auxiliary supervision from retargeted human hand pseudo-labels, weighted by confidence and masked by validity. Arm targets require separate supervision.

At 15~Hz, ACT takes three $240\times320$ RGB views, robot proprioception, and a projection of the tactile latent from 256 to 512 dimensions. Robot touch is sampled at 30~Hz and causally synchronized. ACT predicts chunks of 60 actions, each with 8 or 13 dimensions. The tactile encoder is frozen for 20,000 steps, then further trained at a lower learning rate. Deployment uses robot observations and the learned policy. The framework can also be extended to other robot policy architectures.

\newcommand{\TVcell}{\shortstack{\texttt{[TO}\\\texttt{VERIFY]}}}
\newcommand{\NMcell}{\shortstack{\texttt{[NOT}\\\texttt{YET}\\\texttt{MEASURED]}}}
\newcommand{\PEcell}{\shortstack{\texttt{[PENDING}\\\texttt{EXPERIMENT]}}}

\definecolor{TACROSSNavy}{HTML}{234A73}
\definecolor{TACROSSHeader}{HTML}{DCE8F3}
\definecolor{TACROSSRobotRow}{HTML}{F2F5F8}
\definecolor{TACROSSRokokoRow}{HTML}{EAF4FB}
\definecolor{TACROSSVisionRow}{HTML}{E9F6EE}
\definecolor{TACROSSBest}{HTML}{C6EFCE}
\definecolor{TACROSSOursBlue}{HTML}{DCEEFF}
\definecolor{TACROSSRule}{HTML}{8EA9C1}

\newcommand{\TACROSSTableOneB}{%
\begin{table*}[!tb]
\vspace{-4pt}\caption{Comparison of Full-Hand Tactile Gloves}
\label{tab:rq1-glove-cost}
\centering
\includegraphics[width=\textwidth]{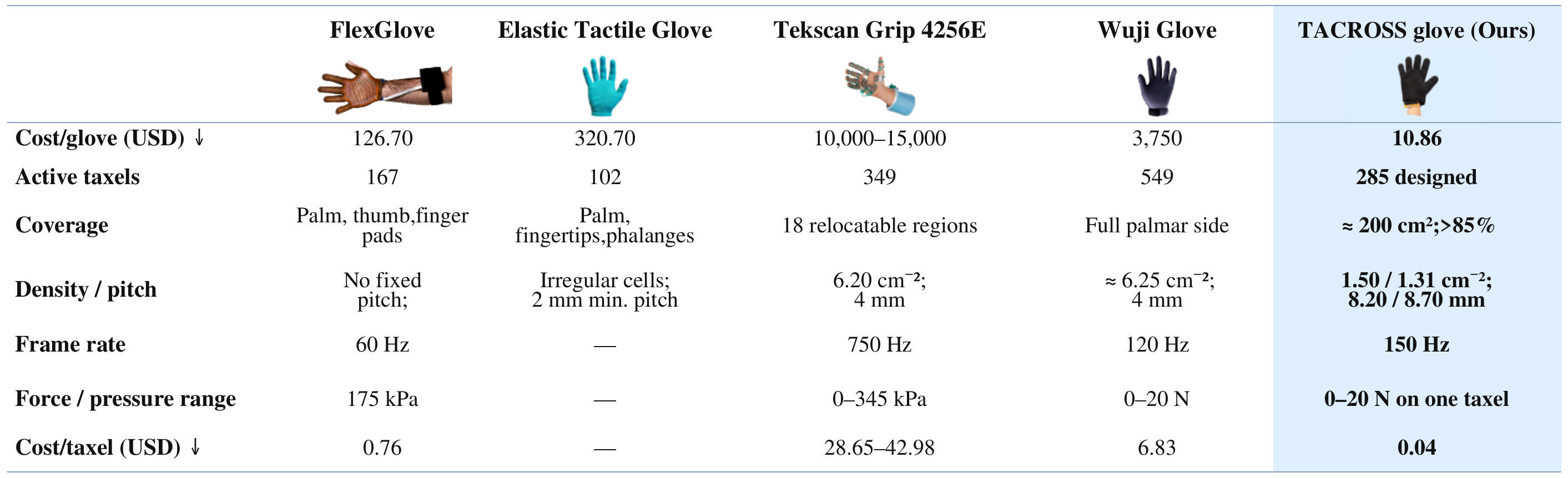}
\par\vspace{3pt}
\begin{minipage}{\textwidth}
\footnotesize
Comparison of full-hand tactile gloves. Specifications are drawn from publications and product documentation. TACROSS values reflect our hardware design and component BOM. The Tekscan Grip 4256E cost (USD~10,000--15,000) and derived cost/taxel are estimates, not a verified manufacturer quotation. Prototype BOMs and commercial system prices have different cost bases.
\end{minipage}
\par

\vspace{-4pt}
\end{table*}%
}

\newcommand{\TACROSSTableTwo}{%
\begin{table}[t]
\vspace{-4pt}
\caption{Reported acquisition equipment costs and recording rates. Rates refer to recorded episodes, while matched valid-demo throughput and end-to-end time accounting remain \toverify.}
\label{tab:rq1-throughput}
\centering
\small
\setlength{\tabcolsep}{3pt}
\renewcommand{\arraystretch}{1.12}
\begin{tabular*}{\columnwidth}{@{}l@{\extracolsep{\fill}}ccc@{}}
\hline
Characteristic & Robot & Ego-R & Ego-V \\
\hline
Equipment cost (USD) & 12{,}000 & 2{,}150 & 520 \\
Recorded episodes/h & 40 & 112 & 140 \\
Rate relative to Robot & $1.0\times$ & $2.8\times$ & $3.5\times$ \\
Robot during human capture & -- & No & No \\
\hline
\end{tabular*}
\par\vspace{2pt}
\begin{minipage}{\columnwidth}
\scriptsize
Ego-R: Rokoko motion capture plus tactile glove, Ego-V: vision-based IK plus tactile glove. Equipment costs describe the acquisition configuration, while robot anchors and robot deployment remain necessary for the mixed-data policies. The USD~10.86 glove BOM is reported separately.
\end{minipage}
\vspace{-4pt}
\end{table}%
}

\newcommand{\TACROSSTableThree}{%
\begin{table*}[t]
\vspace{-4pt}
\caption{\textbf{RQ2: required cross-sensor representation evaluation.}
Retrieval uses group-disjoint held-out splits and same task--object candidate
strata. A match must agree in active fingers, force level, and contact phase.
Every result in this table is \pendingexp.}
\label{tab:rq2-alignment}
\centering
\scriptsize
\setlength{\tabcolsep}{2.1pt}
\renewcommand{\arraystretch}{0.91}
\begin{tabular}{p{0.245\textwidth}cccccccc}
\hline
Method & H$\rightarrow$R R@1 & H$\rightarrow$R R@5 & R$\rightarrow$H R@1 & R$\rightarrow$H R@5 & Contact F1 & Force F1 & Phase F1 & Domain acc. \\
\hline
Raw tactile statistics & \multicolumn{8}{c}{\pendingexp} \\
Binary contact & \multicolumn{8}{c}{\pendingexp} \\
Shared adapter & \multicolumn{8}{c}{\pendingexp} \\
Dual adapters, no cross-domain loss & \multicolumn{8}{c}{\pendingexp} \\
TACROSS, no physical auxiliary losses & \multicolumn{8}{c}{\pendingexp} \\
\textbf{Full TACROSS} & \multicolumn{8}{c}{\pendingexp} \\
\hline
\multicolumn{9}{l}{\textbf{Downstream check at a fixed reduced robot budget of \tbd}}\\
Robot tactile from scratch & \multicolumn{8}{c}{\pendingexp} \\
Human RGB/pose + robot, no human touch & \multicolumn{8}{c}{\pendingexp} \\
Human touch, no cross-sensor alignment & \multicolumn{8}{c}{\pendingexp} \\
Binary-contact alignment & \multicolumn{8}{c}{\pendingexp} \\
\textbf{Full TACROSS} & \multicolumn{8}{c}{\pendingexp} \\
\hline
\end{tabular}
\vspace{-4pt}
\end{table*}%
}

\section{Experiments}
\label{sec:experiments}

\subsection{Experimental Setup}
\label{sec:exp-setup}

\begin{figure}[!t]
\raggedleft
\includegraphics[width=\columnwidth]{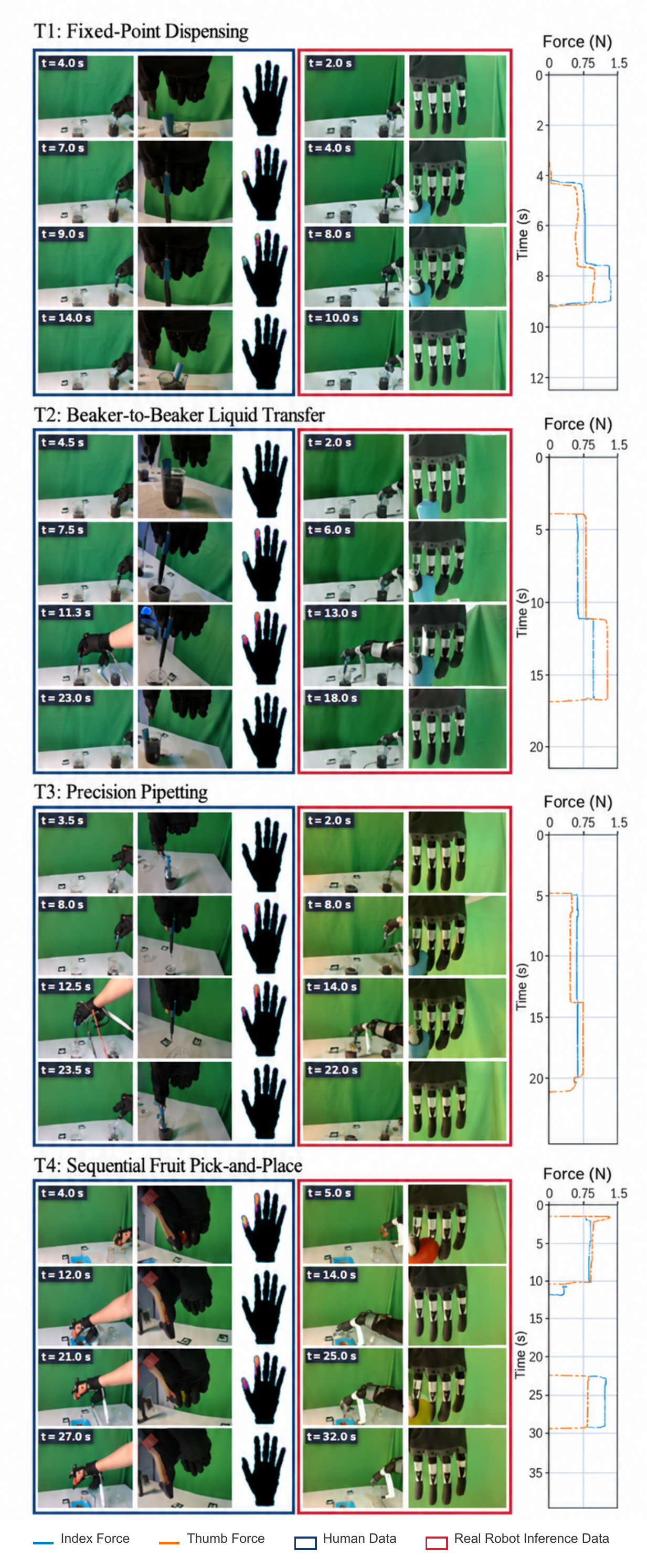}
\caption{Representative human and robot demonstrations for T1--T4 (top to bottom). Each task shows human keyframes (left), tactile maps (middle), robot keyframes (right), and index-finger (blue) and thumb (orange) force traces (far right). Timestamps indicate elapsed time in each demonstration.}
\label{fig:task-design}
\end{figure}

\textbf{Tasks and evaluation.}
We evaluate four contact-rich tasks on real robots: T1 Fixed-Point Dispensing, T2 Beaker-to-Beaker Liquid Transfer, T3 Precision Pipetting, and T4 Sequential Fruit Pick-and-Place. These tasks emphasize accurate dispensing location, stable liquid transfer, dose control, and stable handling of delicate fruit, respectively. A rollout is successful only when it completes all required task stages within the time limit and satisfies the task constraints. For T4, success requires completing the entire prescribed fruit sequence. \hyperref[fig:task-design]{\textcolor{blue!60!black}{Fig.~\ref*{fig:task-design}}} illustrates the four tasks.

For task $k$, success rate is $\mathrm{SR}_k=100s_k/n_k$, where $s_k$ is the number of successful rollouts and $n_k$ is the number of evaluated rollouts. The macro average across the four tasks is $\overline{\mathrm{SR}}=\frac{1}{4}\sum_{k=1}^{4}\mathrm{SR}_k$. We recompute this average from the values for each task before rounding. The separate Sequential Fruit Pick-and-Place experiment on data budgets uses 40 trials per configuration and reports $s_k/n_k$ explicitly.

\textbf{Robot and learning protocol.}
All comparisons use the same visual inputs, interface for robot state, ACT backbone, action space, and closed-loop control settings described in Sec.~\ref{sec:robot-grounding}. Deployment uses robot observations only. Robot demonstrations provide ground-truth action targets, while retargeted human hand targets serve as pseudo-labels for confidence-weighted auxiliary supervision with validity masking. The robot and human training counts are specified with the corresponding policy comparisons below. These configurations are analyzed separately from the larger T4 experiment on data budgets.

\begin{figure}[t]
\vspace{-4pt}
\centering
\includegraphics[width=0.8\columnwidth]{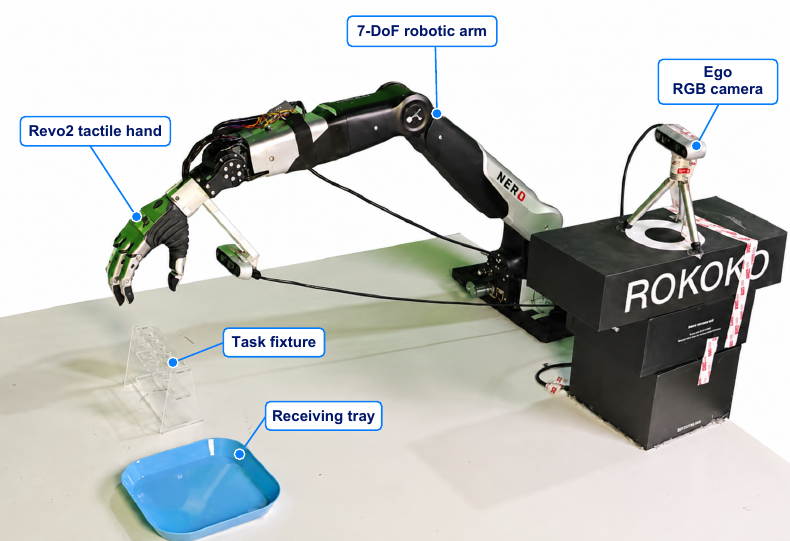}
\caption{Evaluation setup on the robot with the arm, Revo2 tactile hand, RGB cameras, task fixture, and receiving tray.}
\label{fig:robot-experimental-setup}
\vspace{-4pt}
\end{figure}

\subsection{Data Acquisition Cost and Efficiency}
\label{sec:exp-acquisition}

\textbf{Recording throughput and valid data.}
Robot-only teleoperation records approximately 40 episodes/h. Rokoko Ego (Ego-R) and Vision-IK Ego (Ego-V) record approximately 112 and 140 episodes/h, corresponding to $2.8\times$ and $3.5\times$ the robot recording rate. Human capture does not continuously occupy the target robot, enabling additional human contact data to be collected independently of robot availability. \hyperref[tab:acquisition-rebuild]{\textcolor{blue!60!black}{Table~\ref*{tab:acquisition-rebuild}}} also reports corpus size and accepted episodes. Validity rates are recomputed as accepted episodes divided by recorded episodes. Across the three collection configurations, the corpus totals 183.95~h of recorded trajectories, 24,913 recorded episodes, and 22,486 valid episodes, with these totals distinct from subsets used for policy training.

\begin{table}[!htbp]
\centering
\caption{Data Collection Efficiency, System Cost, and Task Performance.}
\label{tab:acquisition-rebuild}
\begingroup
\footnotesize
\setlength{\tabcolsep}{2pt}
\renewcommand{\arraystretch}{1.08}
\begin{tabular*}{\columnwidth}{@{}l@{\extracolsep{\fill}}rrr@{}}
\hline
\textbf{Metric} & \textbf{Robot-only} & \textbf{Ego-R} & \textbf{Ego-V (Ours)} \\
\hline
Total recording duration (h) & 24.86 & 89.49 & \textbf{69.60} \\
Acquisition system cost (USD) $\downarrow$ & 12{,}000 & 2{,}150 & \textbf{520} \\
Recorded demonstrations & 1{,}236 & 12{,}784 & \textbf{10{,}893} \\
Valid demonstration rate (\%) & 86.89 & 91.28 & \textbf{89.44} \\
Valid demonstrations & 1{,}074 & 11{,}669 & \textbf{9{,}743} \\
Collection speedup $\uparrow$ & $1.00\times$ & $2.80\times$ & \textbf{\boldmath$3.50\times$} \\
Average task success (\%) & 92.2 & 93.5 & \textbf{91.5} \\
\hline
\end{tabular*}
\par\vspace{3pt}
\begin{minipage}{\columnwidth}
\scriptsize\raggedright
Robot-only: teleoperation; Ego-R: Rokoko mocap + glove; Ego-V (Ours): vision-based IK + glove; both use robot anchors. Robot:Ego is the demonstration ratio (1:0, 1:5 for the \hyperref[fig:mixing-ratio-ablation]{\textcolor{blue!60!black}{Fig.~\ref*{fig:mixing-ratio-ablation}}} results). Success averages four tasks; speedup is vs.~Robot-only; corpus totals differ from training subsets.
\end{minipage}
\endgroup
\end{table}

\textbf{Equipment investment.}
The full-hand glove has a component BOM of USD~10.86, with 285 physical sensing locations and an output rate of approximately 150~Hz. The current Robot-only, Ego-R, and Ego-V acquisition configurations cost approximately USD~12{,}000, USD~2{,}150, and USD~520. Relative to Robot-only, the listed Ego-R and Ego-V equipment investments are 82.1\% and 95.7\% lower. These costs cover demonstration acquisition equipment. \hyperref[tab:rq1-glove-cost]{\textcolor{blue!60!black}{Table~\ref*{tab:rq1-glove-cost}}} compares glove hardware.

\TACROSSTableOneB

\subsection{Ego Data Gains and Requirements for Robot Demonstrations}
\label{sec:exp-data-gains}

\textbf{Adding Ego data at a fixed robot count.}
\hyperref[tab:multitask-ablation]{\textcolor{blue!60!black}{Table~\ref*{tab:multitask-ablation}}} compares Robot-only with human-augmented training. With the same 30 robot demonstrations, full TACROSS with 150 human demonstrations achieves 91.9\% mean success across four tasks, versus 51.3\% for Robot-only (+40.6 percentage points, pp). Ego data and the transfer pipeline improve task success without additional robot demonstrations. Section~\ref{sec:exp-ablation} examines the contributions of tactile input and alignment.

\begin{table}[!htbp]
\centering
\caption{Ablation Study of Human Tactile Input and Contact-Semantic Alignment}
\label{tab:multitask-ablation}\label{tab:multitask-ablation-ratios}
\begingroup
\fontsize{7}{8.4}\selectfont
\setlength{\tabcolsep}{1.3pt}
\renewcommand{\arraystretch}{1.13}
\begin{tabular*}{\columnwidth}{@{}l@{\extracolsep{\fill}}rrrrrrr@{}}
\hline
\textbf{Characteristic} & \textbf{A} & \multicolumn{2}{c}{\textbf{B}} & \multicolumn{2}{c}{\textbf{C}} & \multicolumn{2}{c}{\textbf{Ours}} \\
Robot:Ego ratio & 1:0 & 1:5 & 1:10 & 1:5 & 1:10 & 1:5 & 1:10 \\
\hline
Robot episodes & 30 & 30 & 16 & 30 & 16 & 30 & 16 \\
Human episodes & 0 & 150 & 160 & 150 & 160 & 150 & 160 \\
\shortstack[l]{Fixed-Point\\Dispensing} & 52.5 & 77.5 & 77.5 & 47.5 & 45.0 & \textbf{95.0} & \textbf{92.5} \\
\shortstack[l]{Beaker-to-Beaker\\Liquid Transfer} & 47.5 & 72.5 & 70.0 & 50.0 & 50.0 & \textbf{92.5} & \textbf{87.5} \\
Precision Pipetting & 50.0 & 67.5 & 65.0 & 47.5 & 45.0 & \textbf{92.5} & \textbf{87.5} \\
\shortstack[l]{Sequential Fruit\\Pick-and-Place} & 55.0 & 75.0 & 72.5 & 52.5 & 47.5 & \textbf{87.5} & \textbf{85.0} \\
Average success & 51.3 & 73.1 & 71.3 & 49.4 & 46.9 & \textbf{91.9} & \textbf{88.1} \\
\hline
\end{tabular*}
\par\vspace{3pt}
\begin{minipage}{\columnwidth}
\scriptsize\raggedright
For B/C/Ours, the left/right subcolumns denote R:H = 1:5/1:10. A: Robot teleoperation only. B: Robot teleoperation + Human Ego without tactile. C: Robot teleoperation + Human Ego without Contact-Semantic Alignment. Ours: Full TACROSS with Contact-Semantic Alignment.
\end{minipage}
\endgroup
\end{table}

\textbf{Increasing human data with 15 fixed robot episodes.}
\hyperref[tab:human-data-budget]{\textcolor{blue!60!black}{Table~\ref*{tab:human-data-budget}}} shows that increasing human demonstrations from 0 to 90 raises four-task mean success from 32.5\% to 85.6\%, demonstrating improved task performance through Ego-data scaling with robot demonstrations fixed at 15.

\begin{table}[!htbp]
\centering
\caption{Scaling human data with 15 fixed robot episodes.}
\label{tab:human-data-budget}
\begingroup
\footnotesize
\setlength{\tabcolsep}{1.5pt}
\renewcommand{\arraystretch}{1.08}
\begin{tabular*}{\columnwidth}{@{}l@{\extracolsep{\fill}}rrrrrr@{}}
\hline
\textbf{Characteristic} & \textbf{A} & \textbf{B} & \textbf{C} & \textbf{D} & \textbf{E} & \textbf{F} \\
\hline
Robot episodes & 15 & 15 & 15 & 15 & \textbf{15} & 105 \\
Human episodes & 0 & 20 & 40 & 60 & \textbf{90} & 0 \\
Human ratio (\%) & 0.0 & 57.1 & 72.7 & 80.0 & \textbf{85.7} & 0.0 \\
Collection cost (USD) & 2.00 & 2.03 & 2.06 & 2.09 & \textbf{2.14} & 14.00 \\
T1 success (\%) & 40.0 & 42.5 & 65.0 & 80.0 & \textbf{90.0} & 95.0 \\
T2 success (\%) & 45.0 & 47.5 & 65.0 & 77.5 & \textbf{85.0} & 87.5 \\
T3 success (\%) & 27.5 & 47.5 & 65.0 & 75.0 & \textbf{82.5} & 82.5 \\
T4 success (\%) & 17.5 & 37.5 & 55.0 & 72.5 & \textbf{85.0} & 87.5 \\
T1 time (min) & 7.3 & 12.1 & 17.0 & 22.5 & \textbf{30.6} & 52.2 \\
T2 time (min) & 15.4 & 22.6 & 29.3 & 37.1 & \textbf{48.1} & 105.4 \\
T3 time (min) & 21.3 & 30.1 & 39.9 & 49.3 & \textbf{63.3} & 148.8 \\
T4 time (min) & 28.9 & 40.8 & 51.5 & 62.8 & \textbf{79.2} & 201.3 \\
\hline
\end{tabular*}
\par\vspace{3pt}
\begin{minipage}{\columnwidth}
\scriptsize\raggedright
A--E use 0, 20, 40, 60, and 90 human episodes, respectively, with 15 robot episodes fixed in each group. F is a robot-only baseline with 105 real-world robot episodes and no human demonstrations. T1--T4: Fixed-Point Dispensing, Beaker-to-Beaker Liquid Transfer, Precision Pipetting, and Sequential Fruit Pick-and-Place.
\end{minipage}
\endgroup
\end{table}

\textbf{Reducing the robot count.}
\hyperref[tab:multitask-ablation-ratios]{\textcolor{blue!60!black}{Table~\ref*{tab:multitask-ablation-ratios}}} compares the 1:5 and 1:10 robot:human allocations. Changing from 30 robot/150 human to 16 robot/160 human demonstrations reduces the robot count by 46.7\%, while mean success decreases from 91.9\% to 88.3\% (-3.6~pp). Ego augmentation reduces robot-demonstration requirements while maintaining high task success.

\begin{figure*}[t]
\centering
\includegraphics[width=0.98\textwidth]{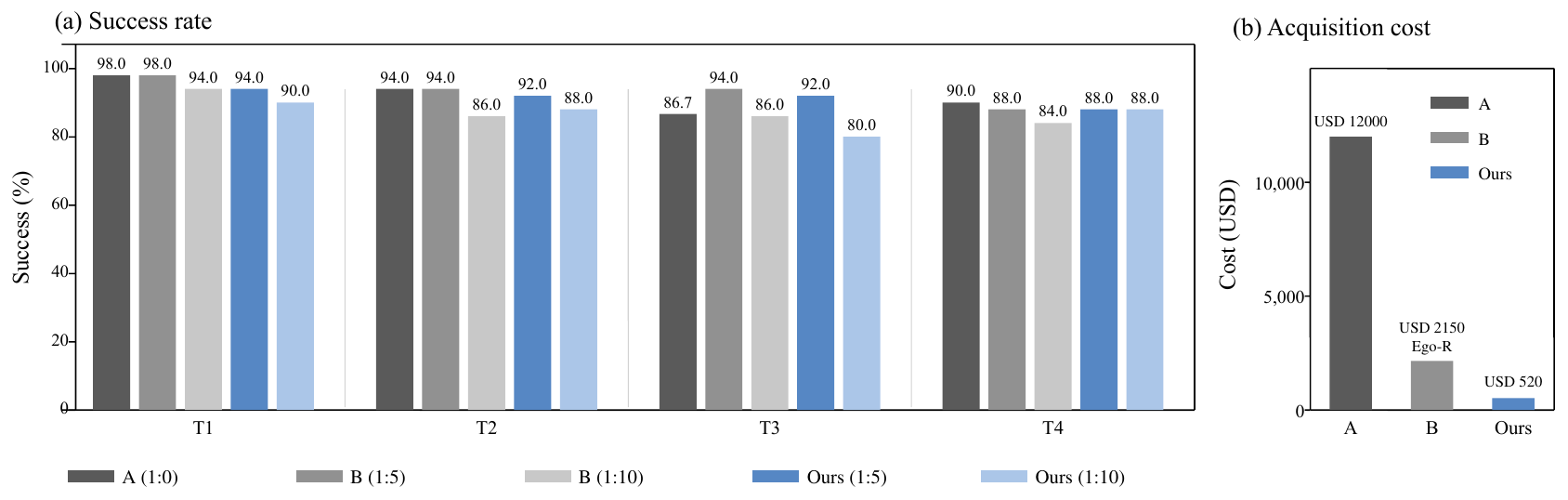}
\caption{Success rates and acquisition equipment costs for different robot--Ego data mixtures. (a) Task success under fixed collection time budgets of 50, 150, 200, and 250 min for T1--T4, respectively. (b) Equipment cost for each acquisition system. A uses robot-only teleoperation, B combines Rokoko motion capture and a tactile glove, and Ours combines vision-based IK and a tactile glove. Both Ego systems use robot demonstrations as anchors. Ratios denote nominal robot:Ego episode counts.}
\label{fig:mixing-ratio-ablation}
\end{figure*}

\textbf{Mixing ratios under fixed budgets for collection time.}
\hyperref[fig:mixing-ratio-ablation]{\textcolor{blue!60!black}{Fig.~\ref*{fig:mixing-ratio-ablation}}} compares Robot-only, Rokoko-based Ego, and vision-IK-based Ego acquisition under fixed per-task collection-time budgets. At the nominal 1:5 ratio, the two Ego systems achieve 93.5\% and 91.5\% macro success, respectively, versus 92.2\% for Robot-only, while at 1:10 they achieve 87.5\% and 86.5\%. The 1:5 mixture offers a better balance between Ego-data use and task success than 1:10.

\textbf{Sequential Fruit Pick-and-Place study of data budgets.}
The separate T4 study evaluates robot--Ego allocations with 40 rollouts per configuration. Robot-only uses 92 robot demonstrations and achieves 38/40 successes (95.0\%). Using 50 robot and 50 Ego demonstrations achieves 37/40 (92.5\%), reducing robot demonstrations by 45.7\% with a 2.5~pp decrease in observed success. At 75\% and 90\% Ego, 25 and 10 robot demonstrations yield 36/40 (90.0\%) and 34/40 (85.0\%), respectively. Ego data can partially replace robot demonstrations on T4, with higher Ego proportions further reducing robot-data requirements but lowering success.

\textbf{Estimated acquisition time.}
For the fruit allocations, acquisition time is estimated as
\begin{equation}
\widehat{T}=N_R\frac{1.500}{\alpha_R}+N_H\frac{60/140}{\alpha_V}\quad\text{min},
\end{equation}
Here, $\alpha_R=1{,}074/1{,}236$ and $\alpha_V=9{,}743/10{,}893$ are the Robot-only and Vision-IK validity rates from \hyperref[tab:acquisition-rebuild]{\textcolor{blue!60!black}{Table~\ref*{tab:acquisition-rebuild}}}. The 50-robot/50-Ego allocation reduces estimated collection time from 158.8 to 110.3~min (30.6\%) relative to 92 robot demonstrations, indicating savings in both acquisition time and robot-data requirements.

\subsection{Ablation Analysis}
\label{sec:exp-ablation}

\textbf{Ego tactile input and contact-semantic alignment.}
Table~\ref{tab:multitask-ablation} compares the full model with variants excluding human tactile input or contact-semantic alignment under two allocations of robot and human training data. With 30 robot and 150 human demonstrations, full TACROSS achieves 91.9\% mean success, compared with 73.1\% without human tactile input and 49.4\% without contact-semantic alignment. With 16 robot and 160 human demonstrations, the corresponding success rates are 88.3\%, 71.3\%, and 46.9\%. Removing human tactile input or contact-semantic alignment reduces success rates in both training settings.

\section{Conclusion}

    We presented TACROSS, an efficient and affordable system for transferring tactile knowledge from humans to robots. It combines tactile acquisition across the hand, alignment of contact semantics across sensors, and policy learning grounded in robot demonstrations with geometric hand retargeting. By combining synchronized human visual, motion, and tactile demonstrations with robot supervision, TACROSS enables faster demonstration collection at lower cost of acquisition equipment while supporting contact-rich dexterous manipulation. This provides a practical way to scale tactile data collection without requiring continuous robot operation. \par The limitation of TACROSS is that the transfer still relies on robot demonstrations and retargeting and calibration specific to each embodiment. In the future, we will extend TACROSS to additional robotic hands and tasks, explore tactile representations across embodiments in a shared latent space, and investigate transfer with smaller budgets for robot demonstrations.

\end{document}